\documentclass{article}

    \PassOptionsToPackage{numbers}{natbib}

    \usepackage[preprint]{neurips_2019}

\usepackage[utf8]{inputenc} 
\usepackage[T1]{fontenc}    
\usepackage{hyperref}       
\usepackage{url}            
\usepackage{booktabs}       
\usepackage{amsfonts}       
\usepackage{nicefrac}       
\usepackage{microtype}      
\usepackage{natbib}
\usepackage{graphicx}
\usepackage{amsmath}
\usepackage{amsfonts}
\usepackage{todonotes}
\usepackage{subcaption}

\title{Task Agnostic Continual Learning via Meta Learning}

\author{%
  Xu He, Jakub Sygnowski, Alexandre Galashov, Andrei A. Rusu, Yee Whye Teh, Razvan Pascanu\\
    DeepMind, London, UK\\
    \texttt{\{hexu, sygi, agalashov, andreirusu, ywteh, razp\}@google.com}\\
}

\begin{document}

\maketitle

\begin{abstract}
While neural networks are powerful function approximators, they suffer from catastrophic forgetting when the data distribution is not stationary.
One particular formalism that studies learning under non-stationary distribution is provided by continual learning, where the non-stationarity is imposed by a sequence of distinct tasks.
Most methods in this space assume, however, the knowledge of task boundaries, and focus on alleviating catastrophic forgetting. In this work, we depart from this view and move the focus 
towards \emph{faster remembering} -- i.e measuring how quickly the network recovers performance rather than measuring the network's performance without any adaptation. 
%
We argue that in many settings this can be more effective and that it opens the door to combining meta-learning and continual learning techniques, leveraging their complementary advantages. 
We propose a framework specific for the scenario where no information about task boundaries or task identity is given. It relies on a separation of concerns into \emph{what} task is being solved and \emph{how} the task should be solved. This framework is implemented by differentiating task specific parameters from task agnostic parameters, where the latter are optimized in a continual meta learning fashion, without access to multiple tasks at the same time. We showcase this framework in a supervised learning scenario and discuss the implications of the proposed formalism.

\end{abstract}

\section{Introduction}

A common assumption made by many machine learning algorithms is that
the observations in the dataset are independent and identically distributed (i.i.d). However, there are many scenarios where this assumption is violated because the underlying data distribution is non-stationary. For instance, in reinforcement learning (RL), the observations depend on the current policy of the agent, which may change over time. In addition, the environments with which the agent interacts are usually non-stationary. In supervised learning tasks, due to computational or legal reasons, one might be forced to re-train a deployed model only 
on the recently collected data, which might come from a different distribution than that of the previous data. In all these scenarios, blindly assuming i.i.d will not only lead to inefficient learning procedure, but also catastrophic interference \citep{mccloskey1989catastrophic}. 

One research area that addresses this problem is continual learning, where the non-stationarity of data is usually described as a sequence of distinct tasks. A list of desiderata for continual learning \citep{schwarz2018progress} include the ability to not forget, forward positive transfer (learning new tasks faster by leveraging previously acquired knowledge), and backwards positive transfer (improvement on previous tasks because of new skills learned), bounded memory budget regardless the number of tasks and so forth. Since these desiderata are often competing with each other, most continual learning methods aim for some of them instead of all, and to simplify the problem, they usually assume that the task labels or the boundaries between different tasks are known.  

In this work, we aim to develop algorithms that can continually learn a sequence of tasks without knowing their labels or boundaries. Furthermore, we argue that in a more challenging scenario where the tasks are not only different but also conflicting with each other, most existing approaches will fail. To overcome these challenges, we propose a framework that applies meta-learning techniques to continual learning problems, and shift our focus from less forgetting to faster remembering: to rapidly recall a previously learned task, given the right context as a cue.

\section{Problem Statement}
We consider the online learning scenario studied by \cite{Hochreiter2001, vinyals2016matching, nagabandi2018deep}, where at each time step $t$, the network receives an input $x_t$ and gives a prediction $\hat{y}_t:=\hat{f}(x_t, \theta_t)$ using a model $\hat{f}$ parametrised by $\theta_t$. It then receives the ground truth $y_t$, which can be used to adapt its parameters and to improve its performance on future predictions. If the data distribution is non-stationary (e.g., $x_t, y_t$ might be sampled from some task $A$ for a while, then the task switches to $B$), then training on the new data might lead to catastrophic forgetting -- the new parameters $\theta'$ can solve task $B$ but not task $A$ anymore: $\hat{f}(x_t^B; \theta')=y_t^B$, $\hat{f}(x_t^A; \theta')\neq y_t^A$. 

Many continual learning methods were proposed to alleviate the problem of catastrophic forgetting. However, most of these approaches require either the information of task index ($A$ or $B$) or at least the moment when the task switches. Only recently, the continual learning community started to focus on task agnostic methods \citep{zeno2018bayesian, online2019Aljundi}. However, all these methods have the underlying assumption that no matter what tasks it has been learning, at any time $t$, it is possible to find parameters $\theta_t$ that fit all previous observations with high accuracy: $\exists \theta_t \ s.t.\ \forall t'\leq t, \hat{f}(x_{t'}, \theta_t)\approx y_{t'}$. This assumption is, however, not valid when the target $y_t$ depends not only on the observation $x_t$ but also on some hidden task (or context) variable $c_t$: $y_t=f(x_t, c_t)$, a common scenario in partially observable environments \citep{monahan1982state, cassandra1994acting}. In this case, when the context has changed ($c_t\neq c_{t'}$), even if the observation remains the same ($x_t=x_{t'}$), the targets may be different ($y_t\neq y_{t'}$). As a result, it is impossible to find a single parameter vector $\theta_t$ that fits both mappings: $\hat{f}(x_{t}; \theta_t)=y_{t} \implies \hat{f}(x_{t'}; \theta_t)\neq y_{t'}$. It follows that, in this case, catastrophic forgetting cannot be avoided without inferring the task variable $c_t$.

\section{What \& How Framework}
Here we propose a framework for task agnostic continual learning that explicitly infers the current task from some context data $\mathcal{D}^{\text{cxt}}_t$ and predicts targets based on both the inputs $x_t$ and task representations $c_t$. The framework consists of two modules: an encoder or task inference network $\mathcal{F}^{\text{what}}:\mathcal{D}^{\text{cxt}}_t\to c_t$ that predicts the current task representation $c_t$ based on the context data $\mathcal{D}^{\text{cxt}}_t$, and a decoder
$\mathcal{F}^{\text{How}}:c_t \to \hat{f}_{c_t}$ that maps the task representation $c_t$ to a task specific model $\hat{f}_{c_t}:x\to \hat{y}$, which makes predictions conditional on the current task. 

\begin{figure*}[h]
\begin{subfigure}[t]{0.33\textwidth}
\centering
    \includegraphics[scale=0.2]{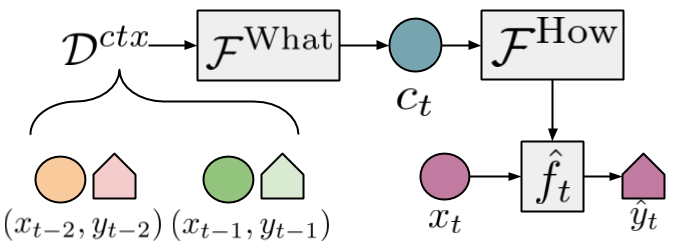}
    \caption{What \& How framework}
    \label{fig:overview}
\end{subfigure}
\begin{subfigure}[t]{0.3\textwidth}
\centering
    \includegraphics[scale=0.21]{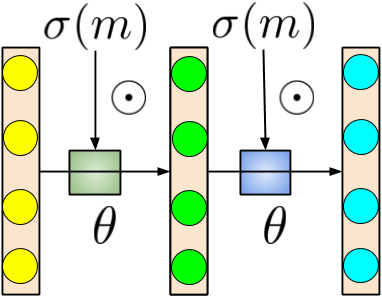}
    \caption{MetaCoG}
    \label{fig:metacog}
\end{subfigure}
\begin{subfigure}[t]{0.3\textwidth}
\centering
    \includegraphics[scale=0.14]{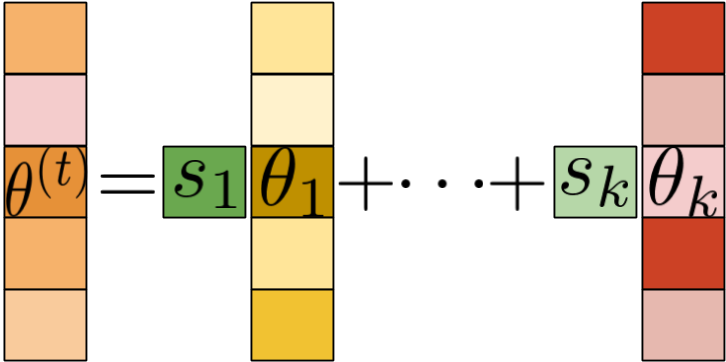}
    \caption{MetaELLA}
    \label{fig:metaella}
\end{subfigure}
\caption{Schematic diagrams of the framework and its instances.}
\end{figure*}

Under this framework, even when the inputs $x_t$ and $x_{t'}$ are the same, the predictions $\hat{y}_{t}$ and $\hat{y}_{t'}$ can differ from each other depending on the contexts. In this work, we choose the recent $k$ observations $\{(x_{t-k}, y_{t-k}),\cdots (x_{t-1}, y_{t-1})\}$ as the context dataset $\mathcal{D}^{\text{cxt}}$. This choice is reasonable in an environment where $c_t$ is piece-wise stationary or changes smoothly. An overview of this framework is illustrated in Figure \ref{fig:overview}.

\subsection{Meta Learning as Task Inference} \label{MetaL}

A similar separation of concern can be found in the meta-learning literature. 
In fact, many recently proposed meta-learning methods can be seen as instances of this framework.
For example, Conditional Neural Processes (CNP) \citep{garnelo2018conditional} embed the observation and target pairs in context data $(x_i, y_i)\in \mathcal{D}^{\text{cxt}}_t$ by an encoder $r_i=h(x_i, y_i;\theta_h)$. The embeddings are then aggregated by a commutative operation $\oplus$ (such as the mean operation) to obtain a single embedding of the context: $r_t= \mathcal{F}^{\text{What}}(\mathcal{D}^{\text{cxt}}_t;\theta_h):= \bigoplus_{x_i, y_i \in \mathcal{D}^{\text{cxt}}_t} h(x_i, y_i; \theta_h)$. At inference time, the context embedding is passed as an additional input to a decoder $g$ to produce the conditional outputs: $\mathcal{F}^{\text{How}}(r_t):= g(\cdot, r_t; \theta_g)$.

 Model-Agnostic Meta-Learning (MAML) \citep{finn2017model} infers the current task by applying one or a few steps of gradient descent on the context data $\mathcal{D}^{\text{cxt}}_t$. The resulting task-specific parameters can be considered a high-dimensional representation of the current task returned by a What encoder: $\theta_t=\mathcal{F}^{\text{What}}(\mathcal{D}^{\text{cxt}}_t;\theta^{\text{init}}):=\theta_t^{\text{init}}-\lambda^{\text{in}}\nabla_{\theta} \mathcal{L}^{\text{in}}(\hat{f}(\cdot; \theta), \mathcal{D}^{\text{cxt}}_t)$, where $\theta_t^{\text{init}}$ are meta parameters, and $\lambda^{\text{in}}$ the inner loop learning rate. The How decoder of MAML returns the task specific model by simply parametrizing $\hat{f}$ with $\theta_t$: $\mathcal{F}^{\text{How}}(\theta_t):=\hat{f}(\cdot; \theta_t)$.

 \cite{rusu2018metalearning} proposed Latent Embedding Optimization (LEO) which combines the encoder/decoder structure with the idea of inner loop fine-tuning from MAML. The latent task embedding $z_t$ is first sampled from a Gaussian distribution $\mathcal{N}(\mu_t^e, diag({\sigma^e_t}^2))$ whose mean $\mu_t^e$ and variance ${\sigma^e_t}^2$ are generated by averaging the outputs of a relation network:
 $\mu_t^e, \sigma^e_t=\frac{1}{|\mathcal{D}^\text{cxt}|^2}\sum_{x_i\in\mathcal{D}^\text{cxt}}\sum_{x_j\in\mathcal{D}^\text{cxt}}g_r(g_e(x_i), g_e(x_j))$, where $g_r(\cdot)$ is a relation network and $g_e(\cdot)$ is an encoder. Task-dependent weights can then be sampled from a decoder $g_d(\cdot)$: $w_t\sim \mathcal{N}(\mu_t^d, diag({\sigma^d_t}^2))$, where $\mu_t^d, \sigma_t^d=g_d(z_t)$. The final task representation is obtained by a few steps of gradient descent: $z_t'=\mathcal{F}^{\text{What}}(\mathcal{D}^{\text{cxt}}_t):=z_t-\lambda^{\text{in}}\nabla_{z'} \mathcal{L}^{\text{in}}(\hat{f}(\cdot; w_t), \mathcal{D}^{\text{cxt}}_t)$, and the final task specific weights $w_t'$ are decoded from $z'$: $\mathcal{F}^{\text{How}}(z_t')=w_t'\sim \mathcal{N}(\mu_t^{d\prime}, diag({\sigma^{d\prime}_t}^2))$, where $\mu_t^{d\prime}, \sigma_t^{d\prime}=g_d(z_t')$. 
 
 In Fast Context Adaptation via Meta-Learning (CAVIA) \citep{zintgraf2019cavia}, a neural network model $\hat{f}$ takes a context vector $c_t$ as an additional input: $\hat{y}=\hat{f}(x, c_t; \theta)$. The context vector is inferred from context data by a few steps of gradient descent: $c_t=\mathcal{F}^{\text{What}}(\mathcal{D}^{\text{cxt}}_t;\theta):=c^{\text{init}}-\lambda^{\text{in}}\nabla_{c} \mathcal{L}^{\text{in}}(\hat{f}(\cdot, c; \theta), \mathcal{D}^{\text{cxt}}_t)$. Then a context dependent model is returned by the How decoder: $\mathcal{F}^{\text{How}}(c_t):= \hat{f}(\cdot, c_t; \theta)$.
 
 Table \ref{metaTable} in Appendix summarizes how these methods can be seen as instances of the What \& How framework. Under this framework, we can separate the task specific parameters of $\hat{f}$ from the task agnostic parameters of $\mathcal{F}^\text{What}$ and $\mathcal{F}^\text{How}$.
\subsection{Continual Meta Learning}
In order to train these meta learning models, one normally has to sample data from multiple tasks at the same time during training. However, this is not feasible in a continual learning scenario, where tasks are encountered sequentially and only a single task is presented to the agent at any moment. As a result, the meta models (What \& How functions) themselves are prone to catastrophic forgetting. Hence, the second necessary component of our framework is to apply continual learning methods to stabilize the learning of meta parameters. In general, any continual learning method that can be adapted to consolidate memory at every iteration instead of at every task switch can be applied in our framework, such as Online EWC \citep{schwarz2018progress} and Memory Aware Synapses (MAS) \citep{aljundi2018memory}. 
In order to highlight the effect of explicit task inference for task agnostic continual learning, we choose a particular method called Bayesian Gradient Descent (BGD) \citep{zeno2018bayesian} to implement our framework. We show that by applying BGD on the meta-level models ($\mathcal{F}^{\text{What}}$ and $\mathcal{F}^{\text{How}}$), the network can continually learn a sequence of tasks that are impossible to learn when BGD is applied to the bottom-level model $\hat{f}$.

Formally, let $\theta^\text{meta}$ be the vector of meta parameters, (i.e. the parameters of $\mathcal{F}^{\text{What}}$ and $\mathcal{F}^{\text{How}}$, for instance, $\theta^\text{init}$ in MAML). We model its distribution by a factorized Gaussian  $p(\theta^\text{meta})=~\prod_i \mathcal{N}(\theta^\text{meta}_i|\mu_i, \sigma_i)$. Given a context dataset $\mathcal{D}^\text{cxt}_t$ and the current observations $(x_t, y_t)$, the meta loss can be defined as the loss of the task specific model on the current observations: $\mathcal{L}^\text{meta}:=\mathcal{L}(\hat{f}_t(x_t), y_t)$, where $\hat{f}_t=\mathcal{F}^\text{How}\circ\mathcal{F}^\text{What}(\mathcal{D}^\text{cxt}; \theta^\text{meta})$. With the meta loss defined, it is then possible to optimize $\mu, \sigma$ using the BGD update rules derived from the online variational Bayes' rule and a re-parametrization trick ($\theta^\text{meta}_i=\mu_i+\sigma_i\epsilon_i,\ \epsilon_i\sim\mathcal{N}(0,1)$):

\begin{align}\label{murule}
    \mu_i \leftarrow & \mu_i - \eta\sigma_i^2
    \mathbb{E}_\epsilon\big[\frac{\partial \mathcal{L}^\text{meta}}{\partial \theta^\text{meta}_i}\big],& \qquad
    \sigma_i \leftarrow & \sigma_i\sqrt{1+\Big(\frac{1}{2}\sigma_i
    \mathbb{E}_\epsilon\big[\frac{\partial \mathcal{L}^\text{meta}}{\partial \theta^\text{meta}_i}\epsilon_i\big]\Big)^2}- \frac{1}{2}\sigma^2_i
    \mathbb{E}_\epsilon\big[\frac{\partial \mathcal{L}^\text{meta}}{\partial \theta^\text{meta}_i}\epsilon_i\big], 
\end{align}

where $\partial \mathcal{L}^\text{meta}/\partial \theta^\text{meta}_i$ is the gradient of the meta loss $\mathcal{L}^\text{meta}$ with respect to sampled parameters $\theta^\text{meta}_i$ and $\eta$ is a learning rate. The expectations are computed via Monte Carlo method: 
\begin{align}\label{MCgrad}
\mathbb{E}_\epsilon\big[\frac{\partial \mathcal{L}^\text{meta}}{\partial \theta^\text{meta}_i}\big]\approx \frac{1}{K}\sum_{k=1}^K \frac{\partial \mathcal{L}^{\text{meta}}(\theta^{\text{meta}(k)}_i)}{\partial \theta^\text{meta}_i}, \qquad  \mathbb{E}_\epsilon\big[\frac{\partial \mathcal{L}^\text{meta}}{\partial \theta^\text{meta}_i}\epsilon_i\big]\approx \frac{1}{K}\sum_{k=1}^K \frac{\partial \mathcal{L}^{\text{meta}}(\theta^{\text{meta}(k)}_i)}{\partial \theta^\text{meta}_i}\epsilon_i^{(k)}
\end{align}

 An intuitive interpretation of BGD learning rules is that weights $\mu_i$ with smaller uncertainty $\sigma_i$ are more important for the knowledge accumulated so far, thus they should change slower in the future in order to preserve the learned skills.

\subsection{Instantiation of the Framework}\label{instances}
Using the What \& How framework, one can compose arbitrarily many continual meta learning methods. To show that this framework is independent from a particular implementation, we propose two such instances by adapting previous continual learning methods to this meta learning framework. 

\paragraph{MetaCoG}
Context-dependent gating of sub-spaces \citep{he2018overcoming}, parameters \citep{mallya2018packnet} or units \citep{serra2018overcoming} of a single network have proven effective at alleviating catastrophic forgetting. Recently, \cite{MasseE10467} showed that combining context dependent gating with a synaptic stabilization method can achieve even better performance than using either method alone. Therefore, we explore the use of context dependent masks as our task representations, and define the task specific model as the sub-network selected by these masks.

At every time step $t$, we infer the latent masks $m_t$ based on the context dataset $\mathcal{D}_t^{\text{cxt}}$ by one or a few steps of gradient descent of an inner loop loss function $\mathcal{L}^{\text{in}}$ with respect to $m$:
\begin{align}
    m_t:= \mathcal{F}^\text{What}(\mathcal{D}_t^{\text{cxt}};\theta) = m^\text{init}-\lambda^{\text{in}} \cdot \nabla_m  \mathcal{L}^{\text{in}}(\hat{f}(\cdot\,; \theta \odot \sigma(m)), \mathcal{D}_t^{\text{cxt}}),
\end{align}

where $m^\text{init}$ is a fixed initial value of the mask variables, $\sigma(\cdot)$ is an element-wise sigmoid function to ensure that the masks are in $[0,1]$, and $\odot$ is element-wise multiplication. In general,  $\mathcal{L}^{\text{in}}$ can be any objective function. For instance, for a regression task, one can use a mean squared error with an $L_1$ regularization that enforces sparsity of $\sigma(m)$:
\begin{align}
\mathcal{L}^{\text{in}}(\hat{f}(\cdot\,; \theta \odot \sigma(m)), \mathcal{D}_t^{\text{cxt}}):=\sum_{x_i, y_i\in \mathcal{D}_t^{\text{cxt}}}(\hat{f}(x_i;\theta \odot \sigma(m))-y_i)^2+\gamma ||\sigma(m)||_1
\end{align}
The resulting masks $m_t$ are then used to gate the base network parameters $\theta_t$ in order to make a context-dependent prediction: $\hat{y}_{t} = \hat{f}(x_t;\theta_t\odot \sigma(m_t))$. Once the ground truth $y_t$ is revealed, we can define the meta loss as the loss of the masked network on the current data: $\mathcal{L}^\text{meta}(\hat{f}(\cdot\,;\theta\odot \sigma(m_t)),\{(x_{t}, y_{t})\})$ and optimize the distribution $q(\theta|\mu, \sigma)$ of task agnostic meta variable $\theta$ by BGD.

The intuition here is that the parameters of the base network should allow fast adaptations of the masks $m_t$. Since the context-dependent gating mechanism is trained in a meta-learning fashion, we call this particular instance of our framework Meta Context-dependent Gating (MetaCoG). We note that while we draw our inspiration from the idea of selecting a subnetwork using the masks $m_t$, in the formulated algorithm $m_t$ rather plays the role of modulating the parameters (i.e. in practice we noticed that entries of $m_t$ do not necessarily converge to $0$ or $1$).  

Note that the inner loop loss $\mathcal{L}^\text{in}$ used to infer the context variable $m_t$ does not have to be the same as the meta loss $\mathcal{L}^\text{meta}$. In fact, one can choose an auxiliary loss function for $\mathcal{L}^\text{in}$ as long as it is informative about the current task.

\paragraph{MetaELLA}
The second instance of the framework is based on the GO-MTL model \citep{kumar2012learning} and the Efficient Lifelong Learning Algorithm (ELLA) \citep{ruvolo2013ella}. In a multitask learning setting, ELLA tries to solve each task with a task specific parameter vector $\theta^{(t)}$ by linearly combining a shared dictionary of $k$ latent model components $L\in\mathbb{R}^{d\times k}$ using a task-specific coefficient vector $s^{(t)}\in\mathbb{R}^k$: $\theta^{(t)}:=Ls^{(t)}$, where $L$ is learned by minimizing the objective function 
\begin{align}
    \label{ellaloss}
    \mathcal{L}^{\text{ella}}(L)=\frac{1}{T}\sum_{t=1}^T \min_{s^{(t)}}\Big\{\frac{1}{n^{(t)}}\sum_{i=1}^{n^{(t)}}\mathcal{L}\big(\hat{f}(x_i^{(t)}; Ls^{(t)}), y_i^{(t)}\big)+\mu||s^{(t)}||_1\Big\}+\lambda||L||_F^2
\end{align}

Instead of directly optimizing $\mathcal{L}^{\text{ella}}(L)$, we adapt ELLA to the What \& How framework by considering $s^{(t)}$ as the task representation returned by a What encoder and $L$ as parameters of a How decoder. The objective $\mathcal{L}^{\text{ella}}$ can then be minimized in a continual meta learning fashion. At time $t$, current task representation $s_t$ is obtained by minimizing the inner loop loss $\mathcal{L}^{\text{in}}(\hat{f}(\cdot; Ls), \mathcal{D}^{\text{cxt}}):=\frac{1}{|\mathcal{D}^{\text{cxt}}|}\sum_{x_i, y_i\in \mathcal{D}^{\text{cxt}}}\mathcal{L}(\hat{f}(x_i; Ls), y_i)+\mu||s||_1$ by one or a few steps of gradient descent from fixed initial value $s^\text{init}$: $s_t:= \mathcal{F}^\text{What}(\mathcal{D}_t^{\text{cxt}}; L)= s^\text{init}-\lambda^{\text{in}} \cdot \nabla_s  \mathcal{L}^{\text{in}}(\hat{f}(\cdot; Ls), \mathcal{D}_t^{\text{cxt}})$. 

Similar to MetaCoG, the parametric distribution $q(L|\mu^L, \sigma^L)=\prod_i \mathcal{N}(L_i|\mu^L_i, \sigma^L_i)$ of the meta variable $L$ can be optimized with respect to the meta loss $\mathcal{L}^\text{meta}(\hat{f}(\cdot; Ls_t),\{(x_{t}, y_{t})\})$ using BGD.

\section{Related Work}
\label{relatedwork}

\textbf{Continual learning} has seen a surge in popularity in the last few years, with multiple approaches being proposed to address the problem of catastrophic forgetting. These approaches can be largely categorized into the following types \citep{parisi2019continual}:
\textit{Rehearsal based methods} rely on solving the multi-task objective, where the performance on all previous tasks is optimized concurrently. 
They focus on techniques to either efficiently store data points from previous tasks \citep{robins1995catastrophic, lopez2017gradient} or to train a generative model to produce pseudo-examples \citep{shin2017continual}. Then the stored and generated data can be used to approximate the losses of previous tasks.
\textit{Structural based methods} exploit modularity to reduce interference, localizing the updates to a subset of weights. \cite{rusu2016progressive} proposed to learn a new module for each task with lateral connection to previous modules. It prevents catastrophic forgetting 
and maximizes forward transfer. 
In \cite{golkar2019continual}, pruning techniques are used to minimize the growth of the model with each observed tasks. 
Finally, \textit{Regularization based methods} draw inspiration from Bayesian learning, and can be seen as utilizing the posterior after learning a sequence of tasks as \emph{a prior} to regularize learning of the new task. These methods differ from each other in how the prior and implicitly the posterior are parametrized and approximated. For instance, Elastic Weight Consolidation (EWC) \citep{kirkpatrick2017overcoming} relies on a Gaussian approximation with a diagonal covariance, estimated using a Laplace approximation. Variational Continual Learning (VCL) \citep{nguyen2017variational} learns directly the parameters of the Gaussian relying on the re-parametrization trick. \cite{ritter2018online} achieved better approximation with block-diagonal covariance. 

While effective at preventing forgetting, the above-mentioned methods either rely on knowledge of task boundaries or require task labels to select a sub-module for adaptation and prediction, hence cannot be directly applied in the task agnostic scenario considered here. To circumvent this issue, \citep{kirkpatrick2017overcoming} used Forget-Me-Not (FMN) \citep{milan2016forget} to detect task boundaries and combined it with EWC to consolidate memory when task switches. However, FMN requires a generative model that computes exact data likelihood, which limits it from scaling to complex tasks. More recently, \cite{online2019Aljundi} proposed a rehearsal-based method to select a finite number of data that are representative of all data seen so far. This method, similar to BGD, assumes that it is possible to learn one model that fits all previous data, neglecting the scenario where different tasks may conflict each other, hence does not allow task-specific adaptations.

\textbf{Meta-learning}, or learning to learn \citep{evolutionary1987sj}, assumes simultaneous access to multiple tasks during meta-training, and focuses on the ability of the agent to quickly learn a new task at meta-testing time. As with continual learning, different families of approaches exist for meta-learning. 
\emph{Memory based methods} \cite{santoro2016meta} rely on a recurrent model (optimizer) such as LSTM to learn a history-dependent update function for the lower-level learner (optimizee). \cite{andrychowicz2016learning} trained an LSTM to replace the stochastic gradient descent algorithm by minimizing the sum of the losses of the optimizees on multiple prior tasks. \cite{ravi2016optimization} use an LSTM-based meta-learner to transform the gradient and loss of the base-learners on every new example to the final updates of the model parameters.
\emph{Metric based methods} learn an embedding space in which other tasks can be solved efficiently. \cite{koch2015siamese} trained siamese networks to tell if two images are similar by converting the distance between their feature embeddings to the probability of whether they are from the same class.
\cite{vinyals2016matching} proposed the matching network to improve the embeddings of a test image and the support images by taking the entire support set as context input. 
The approaches discussed in Section \ref{MetaL} instead belong to the family of \emph{optimization based meta-learning} methods. In this domain, the most relevant work is from \cite{nagabandi2018deep}, where they studied fast adaptation in a non-stationary environment by learning an ensemble of networks, one for each task. Unlike our work, they used MAML for initialization of new networks in the ensemble instead of task inference. A drawback of this approach is that the size of the ensemble grows over time and is unbounded, hence can become memory-consuming when there are many tasks. 

\section{Experiments}
To demonstrate the effectiveness of the proposed framework, we compare BGD and Adam\citep{kingma2014adam} to three instances of this framework on a range of task agnostic continual learning experiments. The first instance is simply applying BGD on the meta variable $\theta^\text{init}$ of MAML instead of on the task specific parameters. We refer to this method as MetaBGD. The other two are MetaCoG and MetaELLA, introduced in Section \ref{instances}. In all experiments, we present $N$ tasks consecutively and each task lasts for $M$ iterations. At every iteration $t$, a batch of $K$ samples $\mathcal{D}_t=\{x_{t,1},\cdots x_{t,K}\}$ from the training set of the current task are presented to the learners, and the context data used for task inference is simply the previous mini-batch with their corresponding targets: $\mathcal{D}^{\text{cxt}}_t=\mathcal{D}_{t-1}\bigcup \{y_{t-1,1},\cdots y_{t-1,K}\}$. At the end of the entire training process, we test the learners' performance on the testing set of every task, given a mini-batch of training data from that task as context data. Since the meta learners take five gradient steps in the inner loop for task inference, we also allow BGD and Adam to take five gradient steps on the context data before testing their performances. We focus on analyzing the main results in this section, experimental details are provided in the Appendix \ref{app:experimentdetails}.

\subsection{Sine Curve Regression}
\begin{figure}[h]
  \centering
  \includegraphics[width=\textwidth]{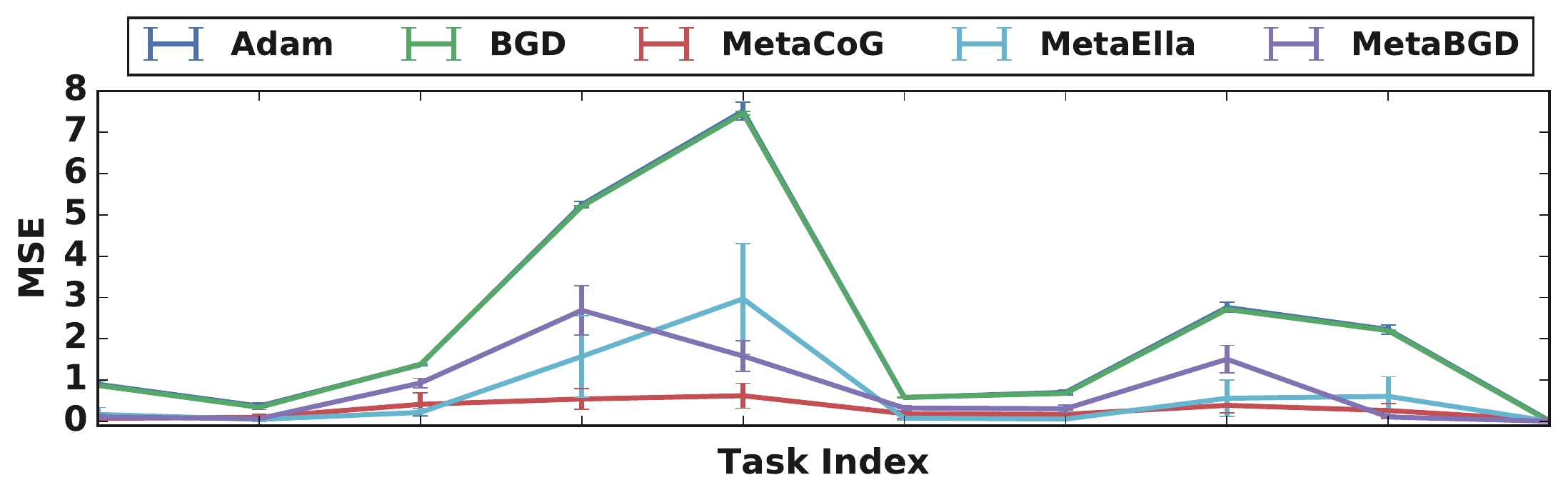}
  \caption{Testing loss per task at the end of the entire learning phase. Task 1 is the first seen task, and task 10 is the last. Lower MSE means better performance.}
  \label{sine_mse}
\end{figure}

\begin{figure}[h]
  \centering
  \includegraphics[width=0.8\textwidth]{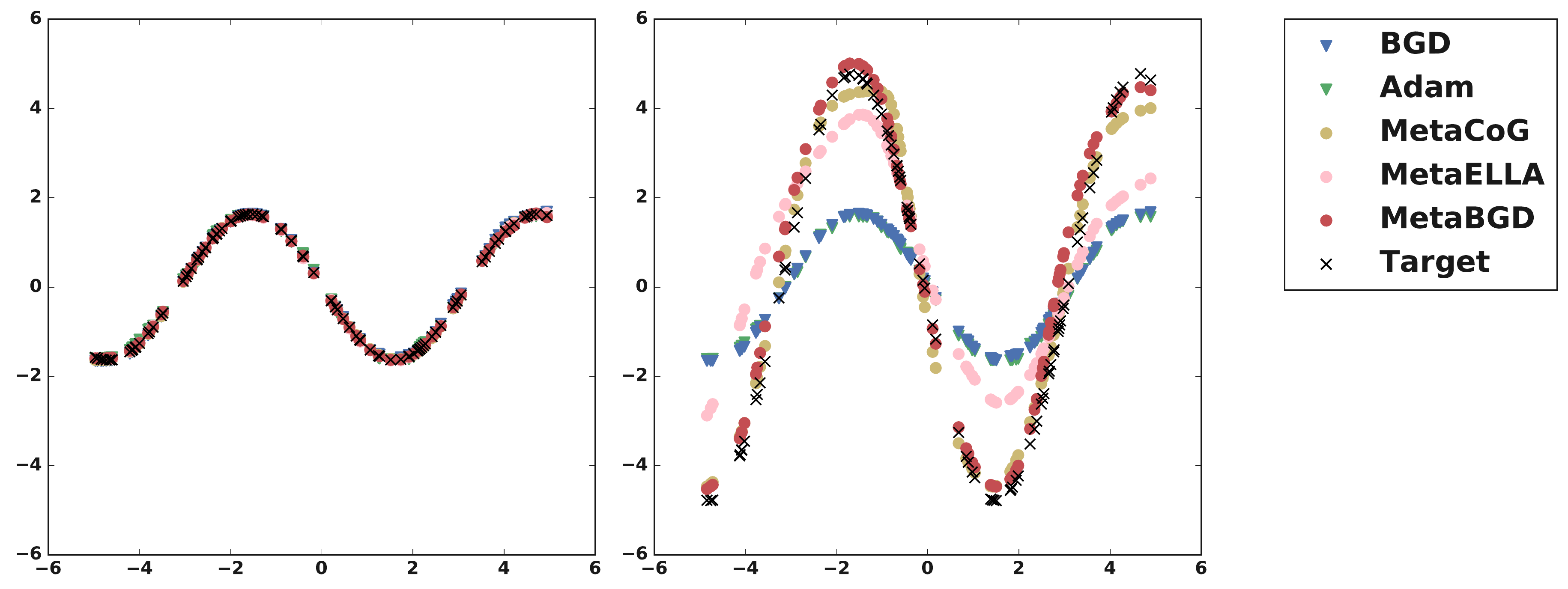}
  \caption{Predictions for the last task (left) and the third task (right) after the entire training process.}
  \label{sine_scatter}
\end{figure}

We start with a regression problem commonly used in meta learning literature, where each task corresponds to a sine curve to be fitted. In this experiment, we randomly generate 10 sine curves and present them sequentially to a 3-layer MLP. Figure \ref{sine_mse} shows the mean squared error (MSE) of each task after the entire training process. Adam and BGD perform significantly worse than the meta learners, even though they have taken the same number of gradient steps on the context data. The reason for this large gap of performance becomes evident by looking at Figure \ref{sine_scatter}, which shows the learners' predictions on testing data of the last task and the third task, given their corresponding context data. All learners can solve the last task almost perfectly, but when the context data of the third task is provided, meta learners can quickly remember it, while BGD and Adam are unable to adapt to the task they have previously learned.

\subsection{Label-Permuted MNIST}

A classical experiment for continual learning is permuted MNIST \citep{goodfellow2013empirical, kirkpatrick2017overcoming}, where a new task is created by shuffling the pixels of all images in MNIST by a fixed permutation. In this experiment, however, we shuffle the classes in the labels instead of the pixels in the images. The reason for this change is to ensure that it is not possible to guess the current task simply based on the images. In this way, we can test whether our framework is able to quickly adapt its behavior according to the current context. Five tasks are created with this method and are presented sequentially to an MLP. 

\begin{figure}[h]
  \centering
  \includegraphics[width=\textwidth]{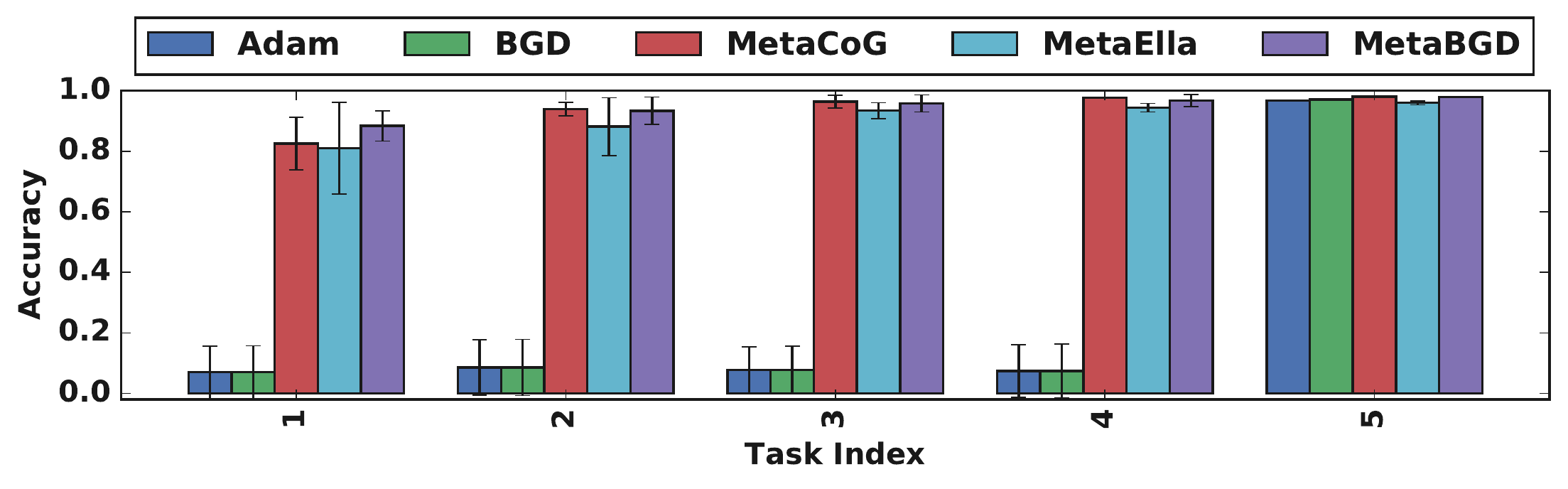}
  \caption{Testing accuracy of different tasks in the label-permuted MNIST experiment at the end of the entire training process.}
  \label{MNIST_acc}
\end{figure}

We test the learners' classification accuracy of each task at the end of the entire learning process, using a mini-batch of training set as context data. As can be seen from Figure \ref{MNIST_acc}, all learners perform well on the last task. However, BGD and Adam have chance-level accuracy on previous tasks due to their incapability of inferring tasks from context data, while the meta learners are able to recall those tasks within 5 inner loop updates on the context data. 

\begin{figure}[h]
  \centering
  \includegraphics[width=\textwidth]{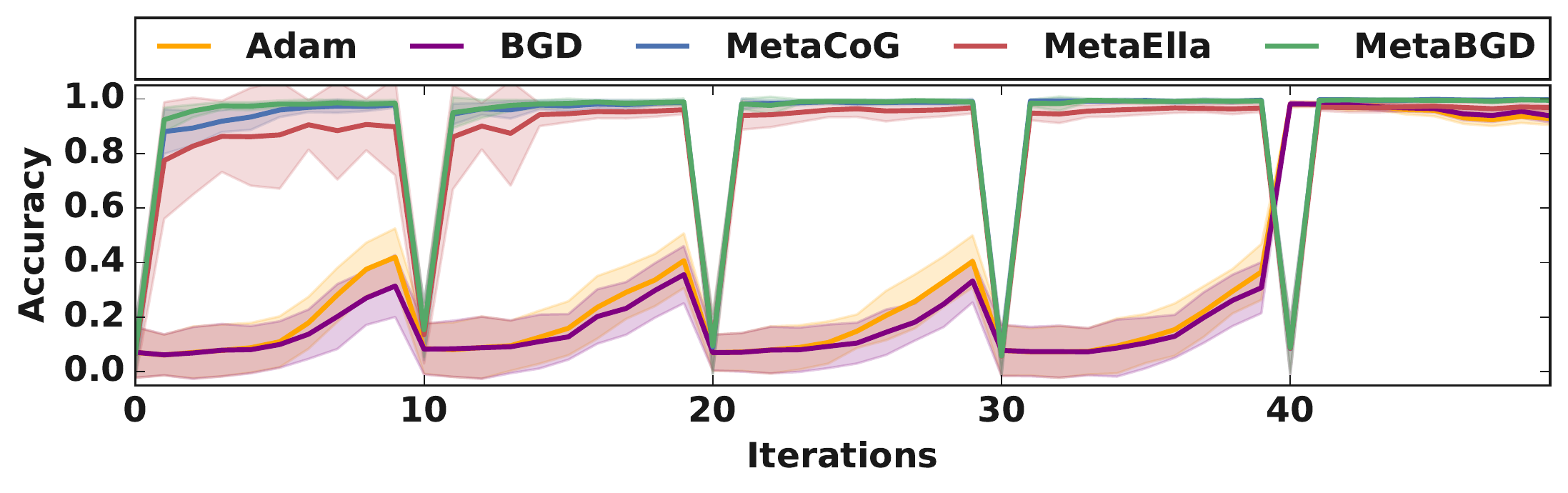}
  \caption{Accuracy curve when the label-permuted MNIST tasks are replayed for 10 iterations after the entire training process. The sudden drops of accuracy are due to task switching, when the context data are still from the previous task. }
  \label{MNIST_review}
\end{figure}

Figure \ref{MNIST_review} displays the accuracy curve when we play the tasks again, for 10 iterations each, after the first training process. The tasks are presented in the same order as they were learned for the first time. It is clear that one iteration after the task changes, when the correct context data is available, the meta learners are able to recall the current task to almost perfection, while Adam and BGD have to re-learn each task from scratch. 

\subsection{Omniglot}
We have seen in previous two experiments that when the task information is hidden from the network, continual learning is impossible without task inference. In this experiment, we show that our framework is favourable even when the task identity is reflected in the inputs. To this end, we test our framework and BGD by sequential learning of handwritten characters from the Omniglot dataset \citep{lake2015human}, which consists of 50 alphabets with various number of characters per alphabet. Considering every alphabet as a task, we present 10 alphabets sequentially to a convolutional neural network and train it to classify 20 characters from each alphabet. 

Most continual learning methods (including BGD) require a multi-head output in order to overcome catastrophic forgetting in this set-up. The idea is to use a separate output layer per task, and to only compute the error on the current head during training and only make predictions from the current head during testing. Therefore, task index has to be available in this case in order to select the correct head. 

\begin{figure}[h]
  \centering
  \includegraphics[width=\textwidth]{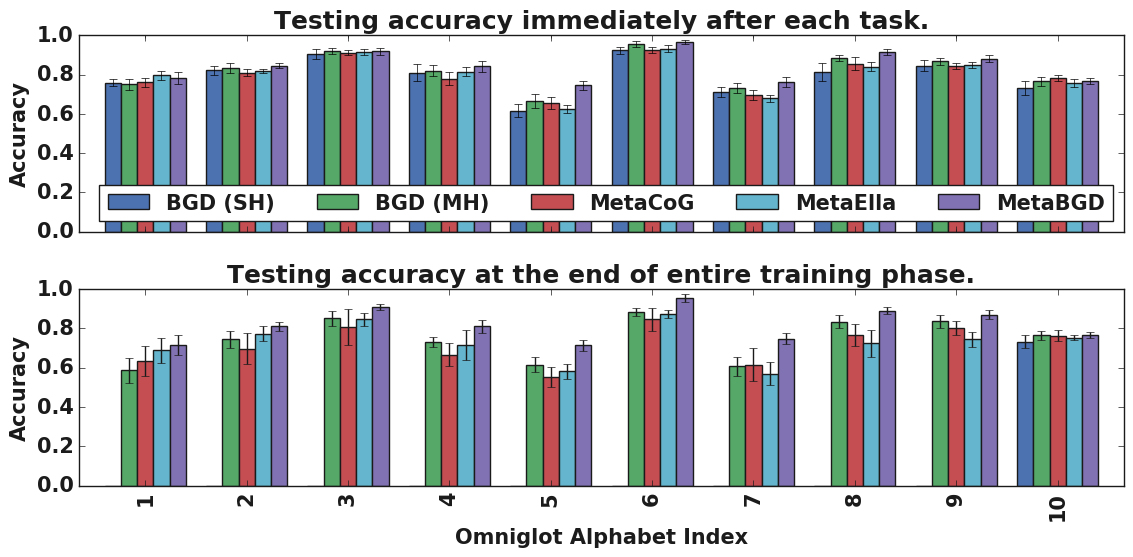}
  \caption{Testing accuracy of the sequential Omniglot task. BGD (MH) uses a multi-head output layer, whereas BGD (SH) and all meta learners use a single-head output layer. In the bottom plot, the accuracy of BGD(SH) are 0 for all tasks except the last one.  }
  \label{Omniglot_acc}
\end{figure}

Unlike these previous works, we evaluate our framework with a single head of 200 output units in this experiment. Figure \ref{Omniglot_acc} summarizes the results of this experiment. For every task, we measure its corresponding testing accuracy twice: once immediately after that task is learned (no forgetting yet), and once after all ten tasks are learned. Our framework with a single head can achieve comparable results as BGD with multiple heads, whereas BGD with a single head completely forgets previous tasks.


\section{Conclusions}
In this work, we showed that when the objective of a learning algorithm depends on both the inputs and context, catastrophic forgetting is inevitable without conditioning the model on the context. A framework that can infer task information explicitly from context data was proposed to resolve this problem. The framework separates the inference process into two components: one for representing \emph{What} task is expected to be solved, and the other for describing \emph{How} to solve the given task. In addition, our framework unifies many meta learning methods and thus establishes a connection between continual learning and meta learning, and leverages the advantages of both.

There are two perspectives of viewing the proposed framework: from the meta learning perspective, our framework addresses the continual meta learning problem by applying continual learning techniques on the meta variables, therefore allowing the meta knowledge to accumulate over an extended period; from the continual learning perspective, our framework addresses the task agnostic continual learning problem by explicitly inferring the task when the task information is not available, and this allows us to shift the focus of continual learning from less forgetting to faster remembering, given the right context. 

For future work, we would like to test this framework for reinforcement learning tasks in partially observable environments, where the optimal policy has to depend on the hidden task or context information.

\bibliographystyle{plainnat}
\bibliography{references}
\newpage
\appendix
\section{Meta Learning as Task Inferences}
\begin{table}[h]
  \caption{Meta learning methods as instances of the What \& How framework}
  \label{metaTable}
  \centering
  \begin{tabular}{llll}
    \toprule
    \cmidrule(r){1-2}
    Methods  & $c_t:=\mathcal{F}^{\text{What}}(\mathcal{D}^{\text{cxt}}_t)$     & $\mathcal{F}^{\text{How}}(c_t)$ \\
    \midrule
    MAML & $\theta_t:=\theta_t^{\text{init}}-\lambda^{\text{in}}\nabla_{\theta} \mathcal{L}^{\text{in}}(\hat{f}(\cdot; \theta), \mathcal{D}^{\text{cxt}}_t)$  & $\hat{f}(\cdot; \theta_t)$    \\
    CNP    & $r_t:= \bigoplus_{x_i, y_i \in \mathcal{D}^{\text{cxt}}_t} h_\theta(x_i, y_i)$ & $g_\theta (\cdot, r_t)$     \\
    LEO   & $z'_t:=z_t-\lambda^{\text{in}}\nabla_{z'} \mathcal{L}^{\text{in}}(\hat{f}(\cdot; w_t), \mathcal{D}^{\text{cxt}}_t)$       & $w_t'\sim \mathcal{N}(\mu_t^{d\prime}(z'_t), diag({\sigma^{d\prime}_t(z'_t)}^2))$  \\
    CAVIA &  $c_t:=c^{\text{init}}-\lambda^{\text{in}}\nabla_{c} \mathcal{L}^{\text{in}}(\hat{f}(\cdot, c; \theta), \mathcal{D}^{\text{cxt}}_t)$               &     $\hat{f}(\cdot, c_t; \theta)$          \\
    \bottomrule
  \end{tabular}
\end{table}
\section{Experiment Details}
\label{app:experimentdetails}
\subsection{Model Configurations}
In all experiments, the number of samples $K$ in Eq. \ref{MCgrad} is set to 10. In MetaCoG, the initial value of masks $m^\text{init}_i$ is 0. In MetaELLA, we use $k=10$ components in the dictionary, and the initial value of latent code $s_i^\text{init}$ is set to $1/k=0.1$. Adam baseline were trained with the default hyperparameters recommended in \cite{kingma2014adam}. The hyperparameters of other methods are tuned by a Bayesian optimization algorithm and are summarized in Table \ref{hyper}. Error bars for all experiments are standard deviations computed from 10 trials with different random seeds. 

\subsection{Sine Curve Regression}
The amplitudes and phases of sine curves are sampled uniformly from $[1.0, 5.0]$ and $[0, \pi]$, respectively. For both training and testing, input data points $x$ are sampled uniformly from $[-5.0, 5.0]$. The size of training and testing sets for each task are 5000 and 100, respectively. Each sine curve is presented for $1000$ iterations, and a mini-batch of 128 data points is provided at every iteration for training. The 3-layer MLP has 50 units with $\tanh(\cdot)$ non-linearity in each hidden layer. 

\subsection{Label-Permuted MNIST}
All tasks are presented for 1000 iterations and the mini-batch size is 128. The network used in this experiment was a MLP with 2 hidden layers of 300 ReLU units. 
\subsection{Omniglot}
We use 20 characters from each alphabet for classification. Out of the 20 images of each character, 15 were used for training and 5 for testing. Each alphabet was trained for 200 epochs with mini-batch size 128. The CNN used in this experiment has two convolutional layers, both with 40 channels and kernel size 5. ReLU and max pooling are applied after each convolution layer, and the output is passed to a fully connected layer of size 300 before the final layer. 

\begin{table}
  \caption{Summary of hyperparameters used for the experiments. $\lambda^\text{in}$ are inner loop learning rates. $\sigma_0$ are initial values for standard deviation of the factorized Gaussian. $\eta$ is the learning rate of the mean in BGD update rule. $\gamma$ is the regularization strength for L1 norm of masks in MetaCoG. $\mu$ is the regularization strength for L1 norm of latent code in MetaELLA.}
  \centering
  \label{hyper}
  \begin{tabular}{rllll}
    \toprule
    \cmidrule(r){1-2}
    Hyperparameters  & Sine Curve & Label-Permuted MNIST & Omniglot \\
    \midrule
    MetaBGD $\lambda^\text{in}$ & 0.0419985  & 0.45 & 0.207496  \\
                    $\sigma_0$ & 0.0368604  & 0.050    & 0.0341916 \\
                    $\eta$   &  5.05646 &  1.0   &   15.8603 \\\hline
    MetaCoG $\lambda^\text{in}$ & 0.849212  & 10.000 &  5.53639   \\
                    $\sigma_0$ & 0.0426860 & 0.034 & 0.0133221    \\
                    $\gamma$ &   1.48236e-6      &  1.000e-5   &  3.04741e-6      \\
                    $\eta$      &  38.6049  &   1.0  &    80.0627 \\\hline
    MetaElla $\lambda^\text{in}$ & 0.0938662  & 0.400 & 0.346027   \\
                      $\sigma_0$ & 0.0298390 & 0.010 & 0.0194483   \\
                          $\mu$ & 0.0216156  &  0.010    & 0.0124128        \\
                          $\eta$      & 42.6035  &     1.0  & 24.7476     \\\hline  
    BGD           $\sigma_0$   &   0.0246160 &  0.060    &    0.0311284   \\    
                  $\eta$      &  20.3049  &    1.0       &    16.2192       \\
    \bottomrule
  \end{tabular}
\end{table}

\end{document}